\documentclass{article}

\usepackage{arxiv}

\usepackage[utf8]{inputenc} % allow utf-8 input
\usepackage[T1]{fontenc}    % use 8-bit T1 fonts
\usepackage{hyperref}       % hyperlinks
\usepackage{url}            % simple URL typesetting
\usepackage{booktabs}       % professional-quality tables
\usepackage{amsfonts}       % blackboard math symbols
\usepackage{nicefrac}       % compact symbols for 1/2, etc.
\usepackage{microtype}      % microtypography
\usepackage{lipsum}
\usepackage{graphicx}
\usepackage{graphicx}%
\usepackage{multirow}%
\usepackage{amsmath,amssymb,amsfonts}%
\usepackage{amsthm}%
\usepackage{mathrsfs}%
\usepackage[title]{appendix}%
\usepackage{xcolor}%
\usepackage{textcomp}%
\usepackage{manyfoot}%
\usepackage{booktabs}%
\usepackage{algorithm}%
\usepackage{algorithmicx}%
\usepackage{algpseudocode}%
\usepackage{listings}%
\usepackage{graphicx}  %%  图片包
\usepackage{subfig}    %% 子图包
\usepackage{float}     %% 浮动个数
\usepackage{sidecap}
\usepackage{amssymb} 
\usepackage[T1]{fontenc}
\usepackage{array}
\usepackage{booktabs}
\usepackage{multirow} 
\usepackage{threeparttable}
\usepackage{hyperref}
\usepackage{fix-cm}
\usepackage{lmodern}
\usepackage{anyfontsize}

\graphicspath{ {./images/} }

\title{Enhanced Cascade Prostate Cancer Classifier in mp-MRI Utilizing Recall Feedback Adaptive Loss and Prior Knowledge-Based Feature Extraction}

\author{
  Kun Luo\thanks{These authors contributed equally to this work.} \\
  School of Automation\\
  Guangdong University of Technology\\
  Guangzhou, Guangdong 510006, China \\
  \texttt{lkkun99668@163.com} \\
  \And
  Bowen Zheng\footnotemark[1] \\
  Department of Urology\\
  Nanfang Hospital, Southern Medical University\\
  Guangzhou, Guangdong 510515, China \\
  \texttt{bowen0762@163.com} \\
  \And
  Shidong Lv \\
  Department of Urology\\
  Nanfang Hospital, Southern Medical University\\
  Guangzhou, Guangdong 510515, China \\
  \texttt{lsd990@smu.edu.cn} \\
  \And
  Jie Tao \\
  School of Automation\\
  Guangdong University of Technology\\
  Guangzhou, Guangdong 510006, China \\
  \texttt{taojiedyxe163.com} \\
  \And
  Qiang Wei \\
  Department of Urology\\
  Nanfang Hospital, Southern Medical University\\
  Guangzhou, Guangdong 510515, China \\
  \texttt{qwei@smu.edu.cn} \\
}

\begin{document}
\maketitle
\begin{abstract}
Prostate cancer is the second most common cancer in males worldwide, and mpMRI is commonly used for diagnosis. However, interpreting mpMRI is challenging and requires expertise from radiologists. This highlights
the urgent need for automated grading in mpMRI. Existing studies lack integration of clinical prior information and suffer from uneven training sample distribution due to prevalence. Therefore, we propose a solution that incorporates prior knowledge, addresses the issue of uneven medical sample distribution, and maintains high interpretability in mpMRI.
Firstly, we introduce Prior Knowledge-Based Feature Extraction, which mathematically models the PI-RADS criteria for prostate cancer as diagnostic information into model training.
Secondly, we propose Adaptive Recall Feedback Loss to address the extremely imbalanced data problem. This method adjusts the training dynamically based on accuracy and recall in the validation set, resulting in high accuracy and recall simultaneously in the testing set.
Thirdly, we design an Enhanced Cascade Prostate Cancer Classiﬁer that classiﬁes prostate cancer into different
levels in an interpretable way, which reﬁnes the classiﬁcation results and helps with clinical intervention. Our method
is validated through experiments on the PI-CAI dataset and outperforms other methods with a more balanced result
in both accuracy and recall rate. 
\end{abstract}

% keywords can be removed
%\keywords{First keyword \and Second keyword \and More}

\section{Introduction}\label{sec1}

Prostate cancer is the second most common cancer and the fifth leading cause of cancer-related death for men worldwide. It has the highest incidence rate among male tumors in over half of the countries  \cite{https://doi.org/10.3322/caac.21660}. Early diagnosis is a prerequisite for subsequent clinical treatment. Therefore, it is crucial to accurately detect clinically significant prostate cancer (csPCa) to avoid overtreatment and reduce mortality. Prostate biopsy is the gold standard for diagnosing csPCa. Currently, biopsies are mostly guided by transrectal ultrasonography (TRUS). However, the difficulty in accurately identifying suspicious nodules using ultrasound poses a challenge, requiring significant expertise. Additionally, biopsy is an invasive procedure that carries risks such as bleeding, infection, and urinary retention. Therefore, a non-invasive and accurate method for diagnosing csPCa is still needed.

Multiparametric magnetic resonance imaging (mpMRI) has become increasingly popular for diagnosing prostate cancer as it provides both anatomical and functional information. It aids in distinguishing csPCa requiring intervention, minimizing overdiagnosis and overtreatment \cite{Wrnschimmel2023}. However, mpMRI interpretation requires substantial expertise and efforts from radiologists, prompting the urgent need for automatic diagnosis of csPCa to ease interpretation burdens and mitigate treatment risks.
	
With the development of artificial intelligence, deep learning is increasingly being applied to medical imaging and has become one of the most important methods in current medical image analysis \cite{10.1145/3065386}. Many researchers have proposed efficient and mature network architectures such as ResNet \cite{DBLP:journals/corr/HeZRS15}, for tasks such as medical image classification and segmentation. Deep learning automates task-specific information learning, eliminating the need for manual extraction efforts. Although it may sacrifice some medical interpretability, it streamlines processes and yields superior performance in targeted tasks.

\subsection{Related works}
	Recently, there has been a rise in computer-aided diagnosis (CAD) solutions for prostate cancer (PCa) using mpMRI images \cite{diagnostics11060959,jamshidi2023radiomics,10468866,app13021088}. Conventional machine learning methods primarily use image-based approaches or radiomics for binary classification tasks \cite{Brancato2021,cancers12092366}. Although these methods yield satisfactory results, they often struggle with more complex multi-classification problems. Deep learning techniques have been extensively explored to address this issue \cite{8653866,10385606,9090311,8245842}, particularly in the context of multi-classification tasks \cite{10085270}. however, due to the inherent complexity of the multi-classification problem, the performance is not as satisfactory as in binary classification \cite{Singhal2022,Nagpal2019}.
	
	Experience and evidence-based medicine are crucial for medical diagnosis. Exploring the integration of them in analyzing images for downstream tasks remains an ongoing endeavor. Matin Hosseinzadeh incorporate the anatomical segmentation mask as prior knowledge to guide the network's focus on the prostate region and improves classification performance \cite{Hosseinzadeh2022}. Alberto Rossi proposed a new deep learning architecture that enables the comparison of new cases with existing cases of prior diagnosis and increase its accuracy \cite{9288754}. Abhejit Rajagopal improved the classification performance by incorporating prior knowledge of histopathology \cite{10483083}. Their work utilizes prior information to improve performance, but the information they use is simplistic and superficial. Complicated information such as diagnostic criteria and medical judgment is not fully utilized.
	
	The imbalanced sample sizes across diseases and stages may bias the classification towards specific samples.Given the cost of misdiagnosis, prioritizing recall rates for these specific samples is crucial. The loss function is a crucial component in deep learning, and an effective loss function enables the model to identify both positive and negative instances. The conventional cross-entropy loss function faces challenges in multi-classification due to strict data conditions. Tsung-Yi Lin improves it for multi-classification by introducing adjustment factors to handle unequal sample class proportions \cite{lin2018focal}. Junjiao Tian introduced Recall Loss, a modification of conventional cross-entropy loss, which incorporates recall as a dynamic parameter. The loss is dynamically weighted based on its changing recall rate every epoch \cite{tian2021recall}. Although these methods have achieved good results, they may not fully suit prostate cancer ISUP classification, especially in distinguishing fewer samples. This potentially leading to locally optimal solutions where samples are classified into a single class, hindering achievement of high recall rates and accuracy simultaneously.
	
	Diagnosis is a complex process that requires a comprehensive understanding of the disease and the patient's condition. It is often divided into multiple stages, each refining the diagnosis.The detection of csPCa is the classification of ISUP 0-1 and ISUP 2-5, which is considered the simplest task in prostate cancer classification. The classification of ISUP 2-3 versus ISUP 4-5 and ISUP 4 versus ISUP 5 is more complex, which indicate the severity of prostate and different clinical interventions. Currently, there is limited research on these complex tasks. The cascade strategy has been used in prostate cancer imaging to simplify complex problems and improve performance. G.J.S. Litjens trained a linear discriminant classifier that sequentially eliminates benign samples using binary classification cascades, ultimately identifying prostate cancer samples \cite{10.1117/12.2043751}. Lina Zhu achieved good performance in fully automated detection and segmentation of csPCa using a Res-UNet cascade deep learning model trained on ADC and T2WI \cite{10.3389/fonc.2022.958065}. Most of the aforementioned cascade works primarily refine classification categories through cascading. However, we believe that the cascade strategy also serves to refine metrics.
	
	\subsection{Contributions}
	In this study, we propose the modeling and strategy design for prostate cancer ISUP classification by combining the prior knowledge of grading criteria and diagnostic procedures to address the aforementioned shortcomings. Our specific contributions are as follows:
	\begin{itemize}
		\item We propose a novel loss function, Recall Feedback Adaptive Loss (RFAloss), which dynamically adjusts during training to expand the parameter search space and adjust the search direction based on the feedback of recall. This addresses the bias caused by the imbalanced sample distribution. We introduce two dynamic parameters and three hyperparameters, evaluate their roles along with the loss function, and prove that RFAloss achieves high recall rates and maintains relatively high accuracy under suitable hyperparameters.
		\item We introduce a prior knowledge-based feature extraction strategy (F-E) based on the report standards of prostate cancer in mpMRI. We validate the effectiveness of this strategy from both visualization and experimental perspectives. When using the F-E results as additional input, it significantly improves the model's ability to generalize on the test set.
		\item We propose a cascade confidence refinement strategy to improve the diagnostic process for physicians. This strategy allows the classifier to output and fuse results based on clinical practice, resulting in a more balanced confusion matrix even with highly imbalanced samples.
	\end{itemize}

	\section{MATERIALS AND METHODS}
	\begin{figure*}[ht]
		\centering
		\includegraphics[width=1\linewidth]{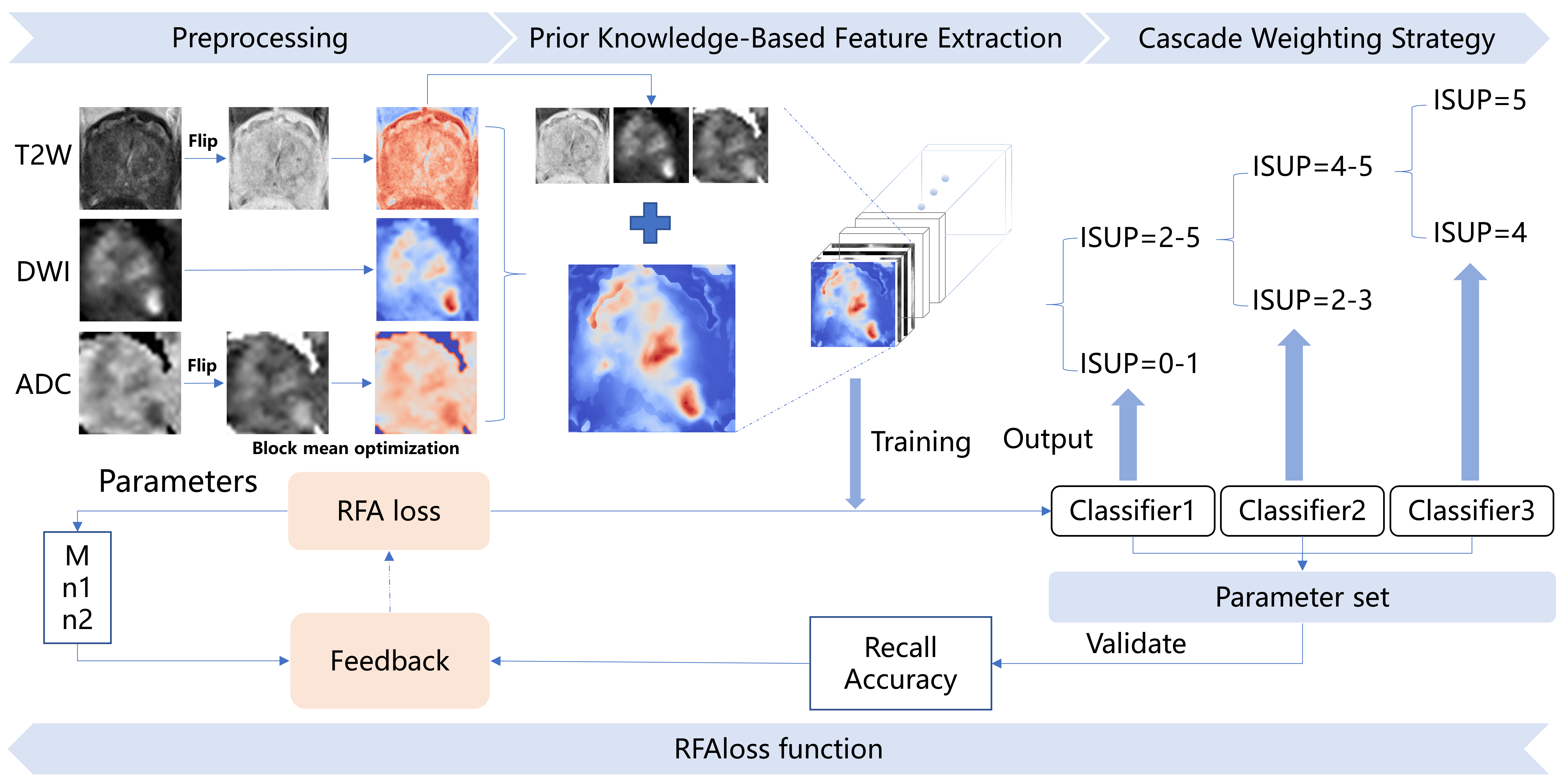}
		\caption{The overall diagram of the proposed method. After preprocessing the T2W, DWI and ADC images, we model the reporting criteria of prostate cancer in mpMRI and design the F-E algorithm to extract features. These results are added as an additional channel for training. Three classifiers are trined to refine the results, and the RFAloss is used to guide the training. The lower part illustrates how the RFAloss works. The accuracy and recall serve as dynamic parameters fed back to the loss function. The hyperparameters $M$, $n_1$, and $n_2$ control the feedback intensity.}
		\label{all_proce}
	\end{figure*}
	
	\subsection{Patient Data and Preprocessing}
	We train and validate our method using the public dataset of PI-CAI Challenge \cite{PICAI_BIAS}, which consists of three sequences: T2WI, ADC, and DWI, along with prostate gland segmentation masks and their ISUP labels for 1499 cases. Considering the data imbalanced of different ISUP stages and the prior knowledge of diagnostic standards for prostate cancer in mpMRI, we preprocess the data as follows:
	
	\subsubsection{Prostate Gland Cropping}
	The effective region for prostate cancer assessment in the original mpMRI images is primarily confined within the prostate gland. Therefore, we crop and resample the data to ensure that the effective training data are within the largest bounding cuboid of the prostate gland. Additionally, we removed a small number of samples that lacked masks. Finally, we resize both width and length to 224 and normalized the pixel values of each point to be between 0 and 255.

	\subsubsection{ADC, T2W Image Signal Flipping}
	Typical prostate cancer shows a hypointense signal in T2WI and ADC images and a hyperintense signal in DWI. To better model image features, we performed signal flipping on the ADC and T2WI sequences. We processed each layer individually, transforming the original lesion from local hypointense to local hyperintense. 
	
	\subsubsection{Block Mean Optimization for Ineffective Region in ADC Images}
	To mitigate the impact of extreme signal levels in ADC and DWI images, we introduce a 2$\times$2 block mean operator for both channels. This operator identifies regions where the mean value exceeds a threshold $\mathcal K_{\text{max}}$ in ADC but remains below a threshold $\mathcal K_{\text{min}}$ in DWI, labeling these areas as ineffective. Subsequently, we invert the pixel values within these regions of the ADC channel to reduce interference from confusing high signals.

	\begin{equation}
		\label{eq:reverse}
		\begin{aligned}
			\text{if } \text{mean}(B_{ADC}) > \mathcal K_{\text{max}}, \text{mean}(B_{DWI}) < \mathcal K_{\text{min}} :\\
			D^{ADC}_k = \max(D^{ADC}_k) - D^{ADC}_{x_B,y_B,k}
		\end{aligned}
	\end{equation}
	
	where $B$ represents the block region, $k$ denotes the k-th layer of the ADC channel, and $x_B$,$y_B$ are the pixel points belonging to the block.

	\subsection{Prior Knowledge-based Feature Extraction (F-E)}
	
	\begin{figure*}[htbp]
		\centering
		\includegraphics[width=1\linewidth]{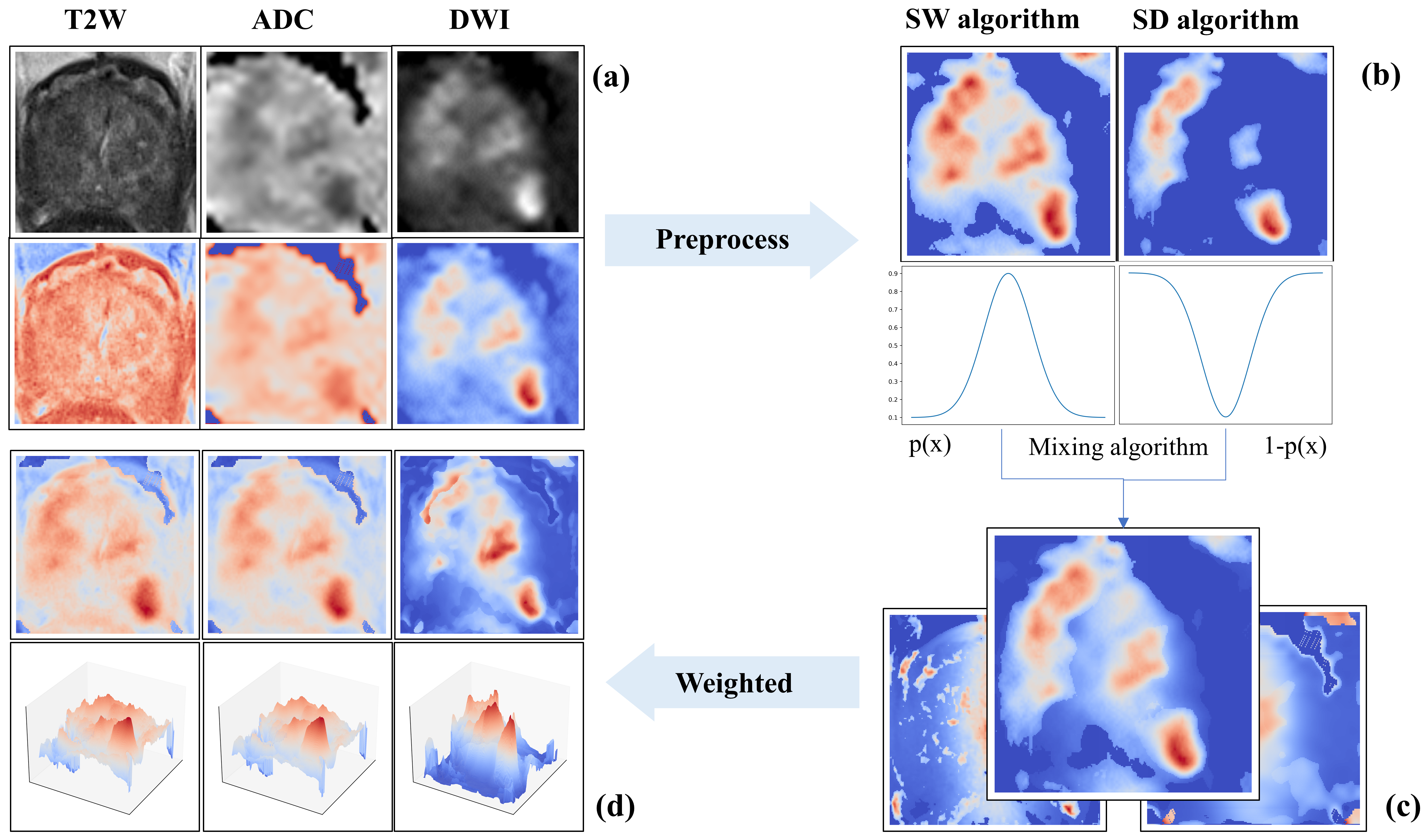}
		\caption{(a) The first row are original images of T2WI, ADC and DWI, and the second row is corresponding preprocessed images. (b) Illustration of the F-E with DWI as an example. (c) The final F-E result was obtained by the weighted addition of (b). (d) The 2d and 3d presentation of (c). The images from left to right show the original picture added directly, the original picture added with weights, and the weighted addition after F-E. A comparison reveals that our feature map significantly enhances regions with high signal across all images, and increase the contrast between peak values and other values.}
		\label{feature_extraction_overview}
	\end{figure*}
	The assessment of prostate cancer in mpMRI is based on the Prostate Imaging Reporting and Data System (PI-RADS) \cite{TURKBEY2019340,WEINREB201616}. Typical prostate cancer shows a hypointense signal in T2WI and ADC images and a hyperintense signal in DWI. The risk of prostate cancer is evaluated by the intensity and the area of the abnormal signal zone. By incorporating this prior knowledge into our model, we improve the generalization ability of the model. This also improve the interpretability of the model, which can help doctors in diagnosis by highlighting areas with a high probability of lesions. 
	
	Quantifying these standards was difficult because signal contrast and local signal range are subjective. To overcome this challenge, we enhanced subjective features by emphasizing important information across different sequences and reducing interference from blurred signals. We quantified local signals by assessing differences in symmetrical positions, where significant intensity discrepancies indicate a high probability of abnormal tissue.Besides, considering that lesions may occur along the gland's axis, we compared the differences between the axial position and surrounding positions. As a result, we refined our theoretical framework and developed the following specific modeling algorithm:
	\subsubsection{Symmetrical Difference Algorithm}
	To quantify the intensity differences in symmetrical positions, we improve upon the classic difference method to highlight signals in the feature area while suppressing signals in non-feature areas. The intensity difference in symmetrical positions is defined as follows:
	
	\begin{equation}
		\label{eq:epsilon}
		\begin{aligned}
			\epsilon_{i,j,k} &= D^N_{i,j,k} - D^N_{i_m-i,j,k}
		\end{aligned}
	\end{equation}
	
	Based on the difference, we propose the Symmetrical Difference (SD) algorithm:
	
	\begin{equation}
		\label{eq:algorithm_D}
		\begin{aligned}
			SD(D^N_{i,j,k}) = 
			\begin{cases}
				\epsilon_{i,j,k}, & \epsilon_{i,j,k} > \phi \\
				0.1, & \epsilon_{i,j,k} \leq \phi
			\end{cases}
		\end{aligned}
	\end{equation}
	
	Where $D^N_{i,j,k}$ represents the pixel at position ($i$, $j$, $k$) in the $\\
	N$-th channel of the mpMRI image, and $\phi$ is a manually set threshold for the symmetric difference. This algorithm efficiently extracts high signal differences at symmetric positions to emphasize the suspicious area.
	
	\subsubsection{Symmetrically weighted Algorithm}
	To capture the differences between the axial line and surrounding positions, we propose a symmetrically weighted algorithm (SW) to quantify the attention to the axial line using a weighted sum and normalization approach. For each row pixel of the image, we designed a weight function that enhances the difference between parts near the axial line and those away from it. Here is the algorithm:
	
	\begin{equation}
		\label{eq:Weight}
		\begin{aligned}
			w(x) = 1 - 0.55 \sin\left(\frac{\pi x}{x_m}\right)
		\end{aligned}
	\end{equation}
	
	\begin{equation}
		\label{eq:SumWeight}
		\begin{aligned}
			SW(D^N_{i,j,k}) = \frac{D^N_{i,j,k}}{\sum_1^{x_m}w(x)D^N_{x,j,k}}
		\end{aligned}
	\end{equation}
	
	Where $w(x)$ is the weight at the horizontal position $x$. This weight helps to keep the attention along the axis low, which means that the weighted sum mainly includes information from areas away from the axial line. Consequently, $SW$ effectively extracts high signals at the axial line position.
	
	\subsubsection{Feature Fusion Strategy}
	
	To accurately represent the results from the SW and SD algorithm, we chose a normal distribution function with a sharper distinction within a specific interval to fuse the two results:

	\begin{equation}
		\label{eq:p(x)}
		\begin{aligned}
			p(x) = \frac{1}{\sqrt{2\pi\sigma^2}} \exp\left(-\frac{(x - \mu)^2}{2\sigma^2}\right)
		\end{aligned}
	\end{equation}
	
	To map $p(x)$ to the interval $(0,1)$, we use X to represent each row and perform the following normalization:
	\begin{equation}
		\label{eq:W(x)}
		\begin{aligned}
			W(x) = \frac{p(x)}{\max(p(X))}
		\end{aligned}
	\end{equation}
	
	Therefore, our feature fusion strategy is as follows:
	
	\begin{equation}
		\label{eq:Mix()}
		\begin{aligned}
			Mix(\zeta ) = W(\zeta )SW(\zeta )+(1-W(\zeta ))SD(\zeta)
			&&\zeta \subset D
		\end{aligned}
	\end{equation}
	
	where D is the pixel points set of mpMRI.The algorithm generates three feature extraction images from the T2WI, ADC, and DWI channels. Since the T2WI emphasizes texture features, there are relatively fewer local signal differences. We fused the T2W, ADC, and DWI channels using a weighted addition with weights of 1:2:2 to create the final feature map.The feature extraction process is illustrated in Figure \ref{feature_extraction_overview}.
	
	\subsection{Recall Feedback Adaptive Loss(RFAloss)}
	
\begin{figure}[htbp]
		\centering
		\includegraphics[width=1\linewidth]{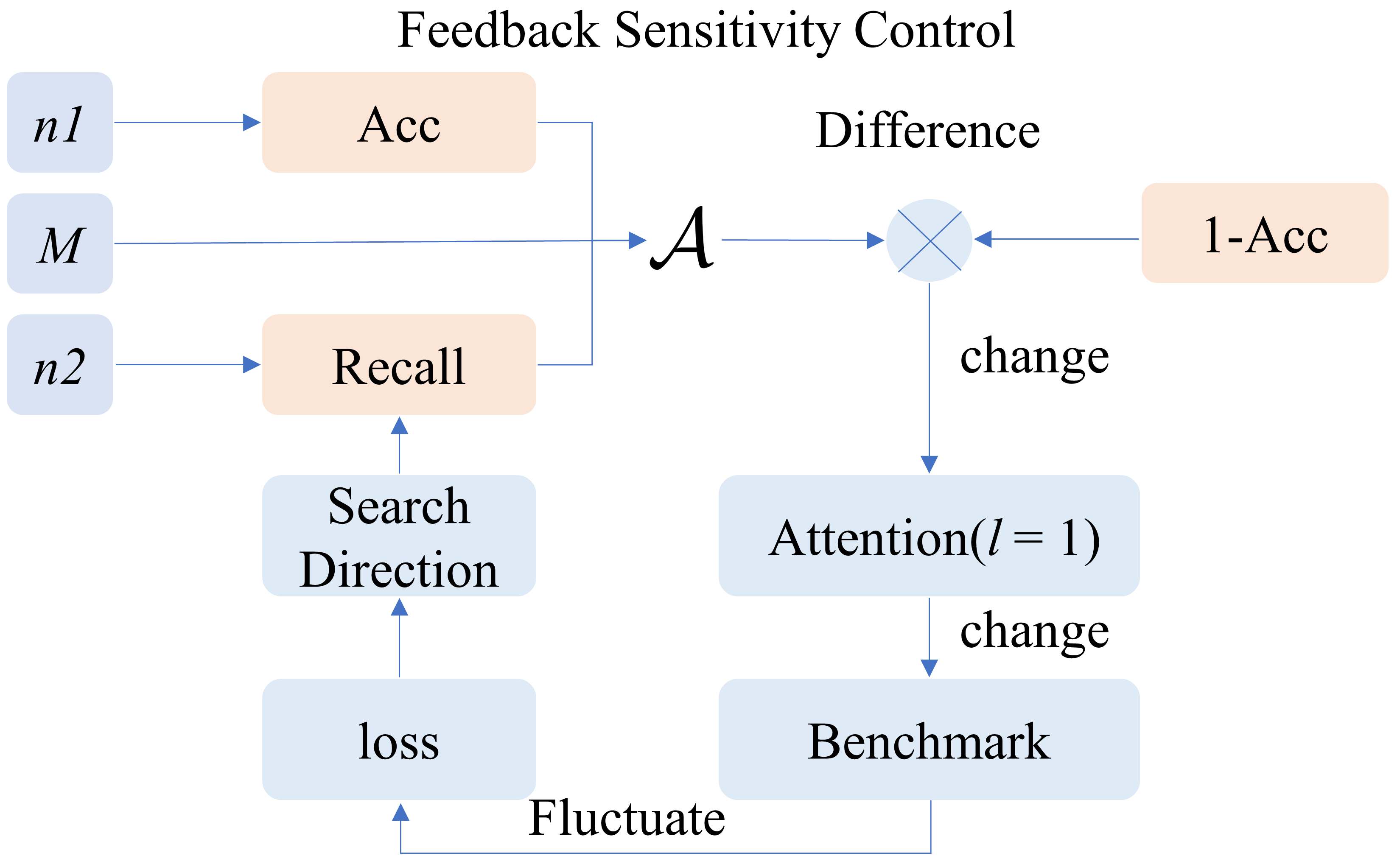}
		\caption{This figure illustrates the mechanism of the Recall Feedback Adaptive loss function, which is controlled by three parameters. Specifically, $n_1$ and $n_2$ determine the feedback sensitivity for accuracy and recall. Together with $M$, they affect the value of the parameter $\mathcal{A}$. The difference between $\mathcal{A}$ and $1-acc$ would change to focus on positive samples, thereby changing the search baseline and causing fluctuations in the loss value. This will finally guide the loss function towards increasing recall.}
		\label{recall示意图}
	\end{figure}

	The training process can be conceptualized as searching for optimal parameters within the parameter space. However, due to the imbalance of data in medical diagnosis, the search process is often biased towards the majority class, leading to suboptimal solutions. Therefore, we constructed a loss function that accurately guides the search direction and widens the search scope. Inspired by control theory, we introduce the Recall Feedback Adaptive Loss:
	\begin{equation}
		\label{eq:RFAloss}
		\begin{aligned}
			\mathcal{L}_{RFA} =
			&\begin{cases}
				\mathcal{A}P_{l=1}-(1-a)\log(P_{l=0}), &  c = 0, \\
				-\mathcal{A}\log(P_{l=1})+(1-a)P_{l=0}, &  c = 1.
			\end{cases}
		\end{aligned}
	\end{equation}
	
	\begin{equation}
		\label{eq:Apart}
		\begin{aligned}
			\mathcal{A} = M\frac{1-a^{n_1}}{{r}^{n_2}}
		\end{aligned}
	\end{equation}
	
	where $P$ represents the probability of the model's output after softmax, $c$ and $l$ denote the predicted label and the true label, respectively. $a$ and $r$ are the accuracy and recall of the validation. $\mathcal{A}$ is used as the adjustment factor. It focuses on positive and negative samples simultaneously and use the accuracy and recall as dynamic feedback to adjusting the $\mathcal{A}$. We control the RFAloss by three hyperparameters $n_1$,$n_2$,$M$ to guide the search direction. Figure \ref{recall示意图} illustrates how the RFAloss works. The design rationale and the mechanism of the RFAloss are detailed in the following sections.
	
	\subsubsection{Setting of Base Framework}
	
	A functional loss function consists of a base structure and functional coefficients. The cross-entropy loss function is commonly used base structure. It is defined as follows:
	
	\begin{equation}
		\label{eq:Cross-Entropy Loss}
		\begin{aligned}
			L_{CE} = -\left(y \cdot \log(P) + (1 - y) \cdot \log(1 - P)\right)
		\end{aligned}
	\end{equation}
	
	where the $y$ and $1-y$ restrict the loss function to focus only on one-hot encoded label. In a high-quality binary classification, the one-hot encoded output label should approach 1 and the non-label should approach 0. The model should be trained to meet this criterion. Therefore, our framework should adhere to the following form:

	\begin{equation}
		\label{eq:base_ori}
		\begin{aligned}
			\mathcal{L}_{base} = f(P_{c=l})+f(P_{c\neq l})
		\end{aligned}
	\end{equation}
	
	where $f(P)$ denotes an operator acting on $P$. We propose distinct output transformation operators to differentiate penalties for correctly and incorrectly classified one-hot encoded labels. For correct classifications, we use the operator $f(P)=-\log(P)$, while for incorrect classifications, we use $f(P)=P$. This make the penalty exhibits a gradual increase as the predicted probability $P$ diminishes towards zero, while maintaining a relatively flat when $P$ approaches one.Therefore, we have finally determined the base framework as:
	
	\begin{equation}
		\label{eq:base}
		\begin{aligned}
			\mathcal{L}_{base} = P_{output\neq label}-\log(P_{output=label})
		\end{aligned}
	\end{equation}
	
	We believe that this framework tends to focus more on correctly classified samples during the early training stages, while still considering other labels. This facilitates the values of correct one-hot encoded label approach 1, while that of incorrect label approach 0.
	
	\subsubsection{Setting of Dynamic Differential Feedback Coefficients}
	
	Misdiagnosis and misclassification of disease severity can be costly for physicians. However, the model often classifies cases into the majority class due to the imbalanced data in clinical. Junjiao Tian's Recall Loss  \cite{tian2021recall} attempted to address this issue by adjusting the recall during training to modify the weights of different classes:
	
	\begin{equation}
		\label{eq:RECALLLOSS}
		\begin{aligned}
			RecallCE = -\sum_{c=1}^C\sum_{n:y_i=c}(1-R_{c,t})\log(p_n,t)
		\end{aligned}
	\end{equation}
	
	$R_{c,t}$ represents the recall of class $c$ at optimization step $t$, and $n:y_i = c$ denotes all samples. Although this loss achieved parameter adjustments along with changes in recall, the change is linear and exhibited low dynamic variability. Furthermore, it lacks sufficient differentiation of disease samples and is susceptible to local optima because it only adjusts one parameter dynamically.
	
	Therefore, we aim to guide the loss function to search towards the recall of the class of interest. We introduce dynamic differential feedback coefficients $\mathcal{A}_0$ and $1-a$. The coefficient $\mathcal{A}_0$ directs the loss function to focus on the cases of interest, while $1-a$ guides it to focus on the cases that are not of interest. The feedback coefficient $\mathcal{A}_0$ is defined as:
	
	\begin{equation}
		\label{eq:A0part}
		\begin{aligned}
			\mathcal{A}_0 = \frac{1-a}{r}
		\end{aligned}
	\end{equation}
	
	where $a$ represents accuracy, $r$ represents recall. When the recall in validation is low, both recall and accuracy are fed back into the training process. This encourages the model to change its search direction, expand its scope, and find the parameter that minimizes loss by maximizing recall and accuracy. Combining Equation \ref{eq:base}, our loss function should be a piecewise function conditioned on the predicted class of the output as follows:
	\begin{equation}
		\label{eq:feedbackloss}
		\begin{aligned}
			\mathcal{L}_{feedback} =
			&\begin{cases}
				\mathcal{A}_0P_{l=1}-(1-a)\log(P_{l=0}), &  c = 0, \\
				-\mathcal{A}_0\log(P_{l=1})+(1-a)P_{l=0}, &  c = 1.
			\end{cases}
		\end{aligned}
	\end{equation}
	
	When the feedback is triggered, the search scope expanded so that the model can escape the local optima. Therefore, instead of using smoothing methods like exponential moving averages, we incorporate average accuracy and recall results per five epochs as feedback parameters in the loss function for adaptive feedback. The effect of RFAloss derives from the difference between $\mathcal{A}_0$ and $1-a$. As $\mathcal{A}_0$ changes dynamically during training, so does the difference between $\mathcal{A}_0$ and $1-a$. After an iteration with a noticeable difference between recall and accuracy, the baseline of the loss function shifts, leading to a significant change in the gradient descent direction. For example, if the recall changes from 1 to 0.5 after an iteration, the attention of the loss function to $P_{l=1}$ increases from one order of magnitude to two orders of magnitude. Consequently, the model prefers to classify $P_{l=1}$ correctly and deviate from its original search direction, leading to both a change in search direction and an expansion of the search scope.
	
	\subsubsection{Setting of Feedback Intensity Hyperparameters}
	
	The proposed loss function in Equation \ref{eq:feedbackloss} already allows for feedback. We further introduce three adjustable hyperparameters for feedback intensity to control the feedback process: $M$, $n_1$, and $n_2$. They are used to improve $\mathcal{A}_0$ into parameter $\mathcal{A}$ as in equation \ref{eq:Apart} , thus allowing control over the loss function.  Our proposed loss aims to make the search direction fluctuate toward the ideal direction. The increase in $n_1$ and $n_2$ exponentially increases the degree of fluctuation. It is important to note that these two parameters, $n_1$ and $n_2$, should not be too large nor too small. This is because the search direction and search scope will restrict each other. Further elaboration on this topic will be provided in the DISCUSSION section.

	\subsection{Cascade Refinement Confidence Strategy}
	In clinical practice, medical diagnosis is a cascade procedure due to the complex nature of diseases. In artificial intelligence, a cascade model can extract complex features from different levels and use the output of one level as input for the next level. This approach may improve the model's performance when dealing with imbalanced data. 
	
	Therefore, we transformed the overall ISUP classification into a cascade classification task. We trained three classifiers (classifier1, classifier2, and classifier3) to perform binary classifications. Classifier1 distinguishes between levels 0-1 and 2-5,and is utilized for diagnosing csPCa, with ISUP levels 0-1 indicating a benign lesion or non-csPCa. while classifier2 separates levels 2-3 from levels 4-5, determines the appropriate clinical interventions for prostate patients, where ISUP levels 2-3 suggest middle-grade csPCa with a relatively positive prognosis and ISUP levels 4-5 indicate high-grade csPCa with an invasion tendency. classifier 3 focuses on classifying level 4 versus level 5, quantifiing the severity of high-grade csPCa, where surgery may be effective for ISUP level 4 patients but not for ISUP level 5 due to increased invasiveness and malignancy. 
	
	To refine the confidence of the final classification, we cascaded the results of the three classifiers. This cascade strategy is illustrated in Figure \ref{all_proce}. We refine the output probabilities of positive classes for each classifier because they work independently. Therefore, we can refine the recall of the subcategory by cascading the multiplication of their relevant recall rates.
	
	\begin{equation}
		\label{eq:RC}
		\begin{aligned}
			R(C(sub-v)) = R(C^n_{l=v}) R(C^{n-1}_{l=sub-v})
			&& sub-v\subset v
		\end{aligned}
	\end{equation}
	
	where $R$ denotes the recall rate of the subcategory, $sub-v$ represents a subset of categories included in the category $v$, and $C^n$ represents the number of classifiers. Through this strategy, we achieved a balanced confusion matrix, even with highly imbalanced samples.

	\section{EXPERIMENTS}
	We conducted multiple experiments to assess the effectiveness of our methods. Since some of the ISUP 0 and 1 labels are generated by artificial intelligence and the significant medical importance of classifying ISUP 2-3 and 4-5, we opted for binary classification (ISUP 2-3 vs. ISUP 4-5) for our Hyperparameter ,ablation and Comparison experiments. Finally, we compared the results of cascade confidence refinement using the optimal RFA loss and feature extraction strategy with those obtained from multi-classification based on cross-entropy.
	
	%We conducted experiments to verify the effects and interpretability of the hyperparameters of RFA loss. Then, we performed ablation experiments under optimal hyperparameters to validate the effects of RFA loss and feature extraction methods.Next, we conduct comparison experiments with three solutions suitable for MRI to illustrate the superiority of our work.Finally, we compared the results of cascade confidence refinement using the optimal RFA loss and feature extraction strategy with those obtained from multi-classification based on cross-entropy, confirming the effectiveness of our work.

	\subsection{Experimental Setup}
	We conducted all the work on NVIDIA 2080Ti. The dataset was divided into training and test sets in a ratio of 9:1. The training set was further split into a training subset and a validation subset in an 8:2 ratio. William's research \cite{10293371} has demonstrated that ResNet has a good classification ability for csPCa. Therefore, We utilized a modified three-dimensional convolutional ResNet101 as the backbone architecture. Adam optimizer was employed with an initial learning rate of 0.0005, which was reduced to 1/10 of the original rate at 100 and 200 epochs. A batch size of 16 was utilized, and iterations were continued until reaching 500 epochs or until significant early convergence occurred.
	
	Taking into account the stochastic nature of the fluctuation search, we save results where accuracy is above 0.7 and recall is above 0.6 as excellent parameters. The set of parameters for the test process is from the excellent parameter set.

	\subsection{Hyperparameter Experiment for RFA Loss}
	
	To evaluate the general effects of various hyperparameters of the loss function, we conducted the following experiments while keeping other variables constant. Five-fold cross-validation was performed on the training set, and the mean of the optimal results was used as the experimental indicator for the group of hyperparameters.
	
	Firstly, we conducted four experiments with \( n_1 \) set to 0.25, 0.5, 0.75, and 1 while keeping \( n_2 \) and \( M \) fixed at 3 and 0.3 respectively. The goal was to amplify the fluctuation of Equation \ref{eq:Apart} by adjusting \( n_1 \).
	
	Next, to assess the influence of \( n_2 \), we set \( n_1 = 1 \) and \( M = 0.5 \). We then varied \( n_2 \) from 1 to 3 and evaluated its effect.
	
	Finally, we conducted an experiment to evaluate the impact of \( M \) on the entire system. We controlled \( n_1 \) and \( n_2 \) at values of 1 and 3 respectively, and performed three experiments with different values of \( M \): 0.3, 0.5, and 0.7.
	
	\subsection{Ablation Experiments}
	To validate the effectiveness of our loss function, we compared it with classical ones such as cross-entropy loss, focal loss, and recall loss. We performed three experiments on training sets with different random seed ,and took the average of the first three best results on the test set as the result of the ablation experiment. This comparison allowed us to verify the feedback effect and final performance of our proposed loss function. Additionally, we assessed the effectiveness of the feature extraction module by integrating it as an additional channel input into different loss functions. Finally, we evaluated the synergistic effect of combining both methods.
	
	\subsection{Comparison experiment}
	To verify the superiority of our work, we compared it with three methods: M3T \cite{9878673}, HiFuse \cite{huo2022hifuse}, and MedViT \cite{Manzari_2023}. M3T combines CNN and Transformer model for 3D medical image classification. HiFuse and MedViT are trained on 2D images and then tested on the patient level. We conducted classification experiment of ISUP 2-3 vs 4-5 using these three schemes on the PI-CAI data set. We did not find related work for ISUP 2-3 vs 4-5 classification,so we refer to the experimental results of Gianluca Carloni performing similar work on the PI-CAI data set \cite{carloni2023causalitydriven}.We used their optimal results as an experimental indicator for the comparison,and they are compared with our scheme on evaluation metrics to illustrate the effectiveness of our method.
	
	\subsection{Evaluation of Cascaded Refinement Confidence Strategy}
	We evaluated the effectiveness of our work by using optimal parameters from ISUP 2-3 and 4-5 classification to perform two binary classification tasks (ISUP 0-1 versus ISUP 2-5, ISUP 4 versus ISUP 5). We compared our proposed method with a baseline six-class ISUP classification based on cross-entropy to assess its efficacy and superiority.
	
	\subsection{Evaluation Metrics}
	We use recall and accuracy (acc) to evaluate classification performance and compute precision for the samples of interest. In hyperparameter experiments, we propose an acc-recall score(ARS) to simultaneously assess the fusion results of recall and precision with equal weights and different weights. For the comprehensive evaluation, we adopt two metrics: ARS score and F2-Score. In the ablation experiments, we use F2-Score and Area Under the Curve (AUC) as evaluation metrics. 
	
	The acc-recall score is the geometric mean of recall and accuracy:
	\begin{equation}
		\label{eq:G}
		\begin{aligned}
			ARS = \sqrt{r \cdot a}
		\end{aligned}
	\end{equation}
	where $r$ represents recall and $a$ represents accuracy.
	
	The F-score is a measure of predictive performance. Positive real factor in the F2-Score is 2 to defines recall as twice as important as precision.  The AUC is the area under the ROC (Receiver Operating Characteristic) curve. The AUC value ranges from 0.5 to 1, where a higher value closer to 1 indicates greater accuracy in detection methods. An AUC of 0.5 suggests low accuracy and no practical value.
	
	To better understand the fluctuation of loss functions and the impact of hyperparameters,  we use visualization methods to depict the descent of training losses, which helps with auxiliary analysis and interpretation.To evaluate the effectiveness of our cascaded refinement confidence strategy, the confusion matrix is used.
	
	\section{RESULTS}
	
	\begin{figure*}[ht]
		\centering
		\subfloat[$n_1$]
		{\includegraphics[width=5cm]{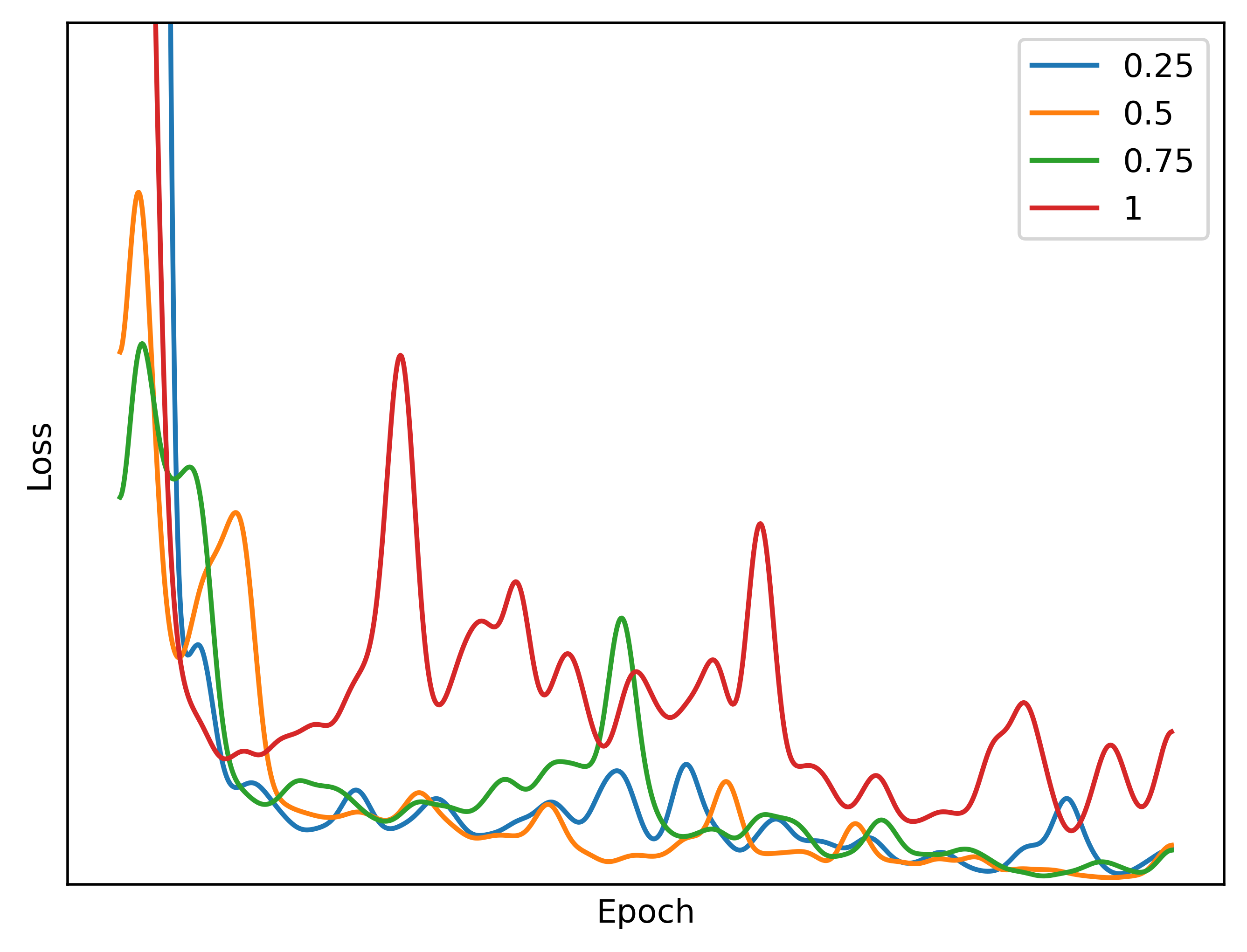}\label{n1pic}}
		% \hspace{-1mm}
		\subfloat[$n_2$]
		{\includegraphics[width=5cm]{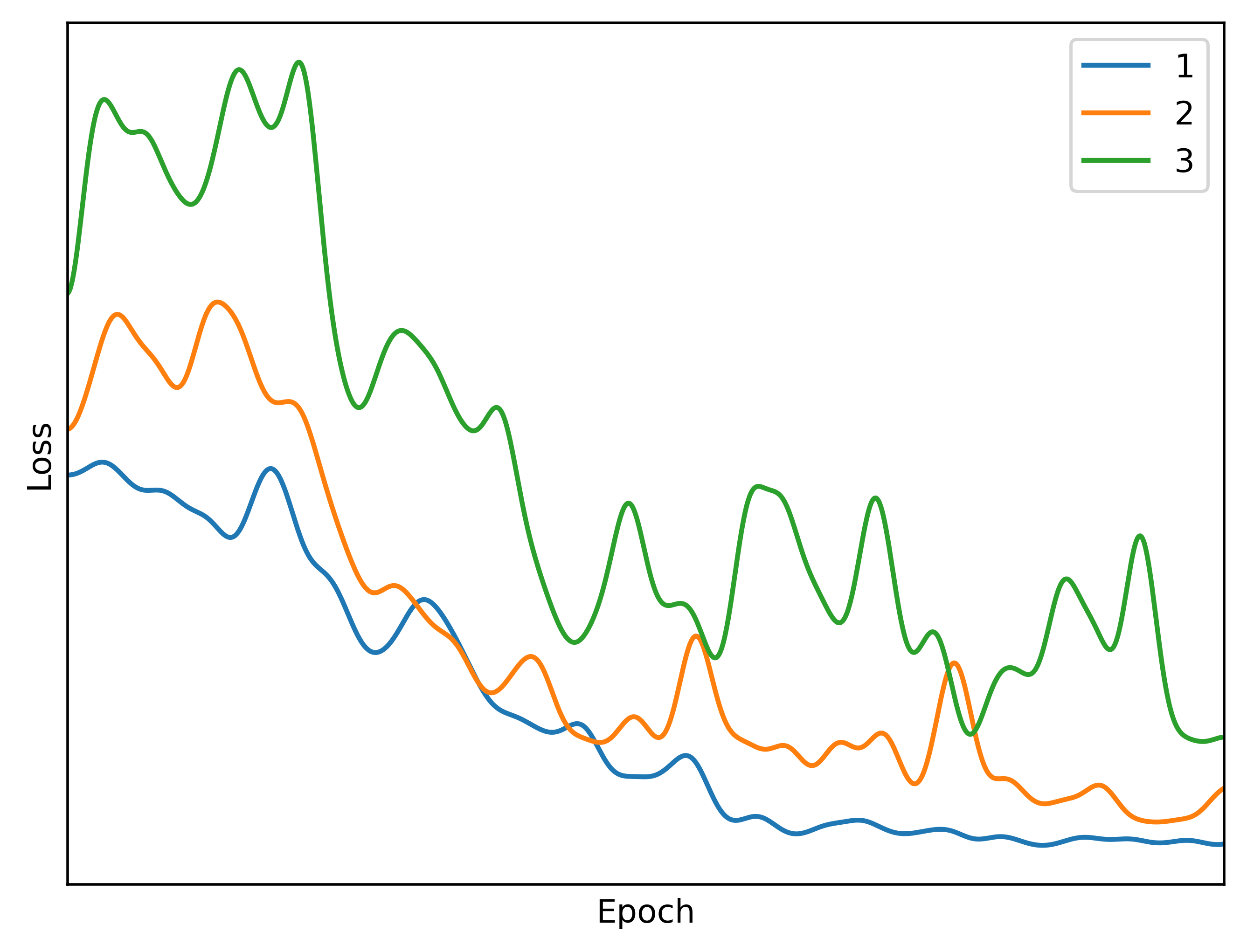}\label{n2pic}}
		% \hspace{-13mm}
		\subfloat[$M$]
		{\includegraphics[width=5cm]{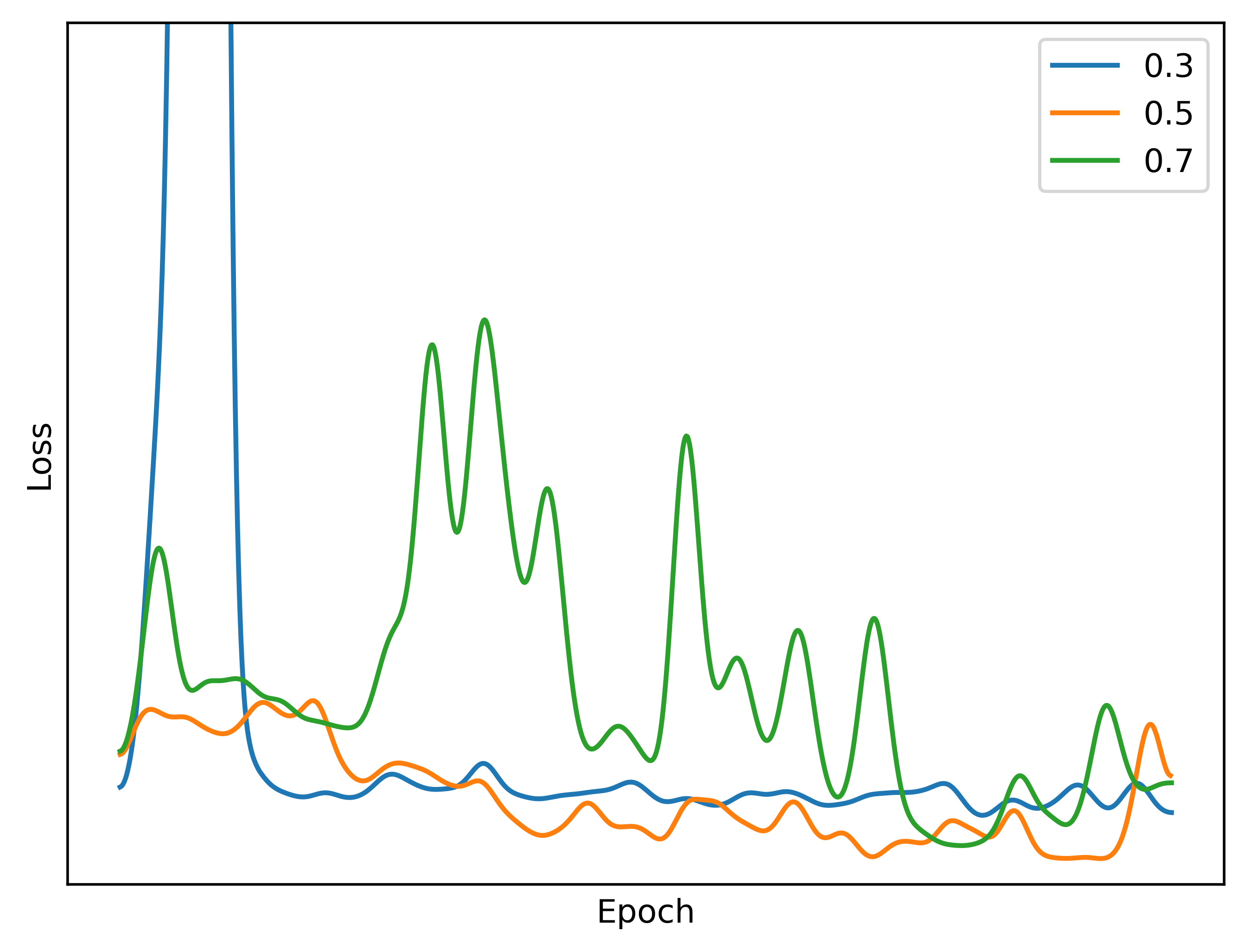}\label{Cpic}}
		\caption{Figures \ref{n1pic}, \ref{n2pic}, and \ref{Cpic} show the training loss curves for hyperparameters $n_1$, $n_2$, and $M$. The curves have been smoothed using Gaussian smoothing. The fluctuation of $n_1$ and $n_2$ decreases as their values decrease, while the fluctuation of $M$ is higher at 0.7 and slightly lower at 0.3 compared to 0.5.}
		\label{pic:control parameter curve}
	\end{figure*}

\begin{table}[htbp]
\caption{Result of Hyperparameter Experiments}\label{table:control_variables}
\centering
\begin{tabular}{@{}lllllllll@{}}

\toprule
~ & $M$ & $n_1$ & $n_2$ & Acc & Prec & Recall & ARS & F2-score  \\ 
\midrule
\multirow{4}{*}{$n_1$}& 0.3 & 0.25 & 3 & 0.688 & 0.449 & 0.505 & 0.589 & 0.493 \\ 
~ & 0.3 & 0.5 & 3 & 0.784 & 0.498 & 0.528 & 0.643 & 0.522 \\ 
~ & 0.3 & 0.75 & 3 & 0.744 & 0.667 & 0.610 & 0.673 & 0.620 \\ 
~ & 0.3 & 1 & 3 & 0.798 & 0.538 & 0.555 & 0.665 & 0.551 \\ 
\midrule
\multirow{3}{*}{$n_2$} & 0.5 & 1 & 1 & 0.700 & 0.460 & 0.390 & 0.522 & 0.402 \\ 
~ & 0.5 & 1 & 2 & 0.754 & 0.543 & 0.628 & 0.688 & 0.609 \\ 
~ & 0.5 & 1 & 3 & 0.709 & 0.317 & 0.334 & 0.486 & 0.330 \\ 
\midrule
\multirow{3}{*}{$M$} & 0.3 & 1 & 3 & 0.798 & 0.538 & 0.555 & 0.665 & 0.551 \\ 
~ & 0.5 & 1 & 3 & 0.709 & 0.317 & 0.334 & 0.486 & 0.330 \\ 
~ & 0.7 & 1 & 3 & 0.641 & 0.424 & 0.341 & 0.468 & 0.355 \\ 
\bottomrule
\end{tabular}
\footnotetext[1]{The CH column represents controlled variables in the experiments on hyperparameters, delineated by a solid line into three sets of experiments.}
\footnotetext[2]{$M$, $n_1$, and $n_2$ denote different hyperparameters, while Prec refer to Precision.}
\end{table}

\begin{table}[htbp]
\centering
\caption{Ablation Experiment Results}\label{table:ablation_experiment}
\begin{tabular}{@{}ccccccc@{}}
\toprule
F-E & Loss & Acc & Prec & Recall & F2-score & AUC  \\ 
\midrule
~ & Focal & 0.792  & 0.476   & 0.389   & 0.404   & 0.647   \\ 
~ & CE & 0.792  & 0.450  & 0.519  & 0.503   & 0.695   \\ 
~ & Recall & 0.281  & 0.281  & 1.000  & 0.661   & 0.500   \\ 
~ & RFA & 0.800  & 0.481  & 0.681  & 0.629   & 0.756   \\
$\surd$ & Focal & 0.825  & 0.398  & 0.704  & 0.610   & 0.782  \\ 
$\surd$ & CE & 0.720  & 0.561  & 0.424 & 0.446   & 0.633  \\ 
$\surd$ & Recall & 0.281 & 0.281  & 1.000  & 0.661   & 0.500   \\ 
$\surd$ & RFA & 0.800  & 0.471  & 0.810   & 0.708    & 0.804    \\ 
\bottomrule
\end{tabular}
\footnotetext[1]{F-E refers to the feature extraction operation, while the Loss column represents the chosen loss function for the ablation experiment. Prec refers to Precision.}
\end{table}

	% \begin{table}[htbp]
	% 	\caption{Comparison Experiment Results}
	% 	\label{table:Comparison_experiment}
	% 	\centering
	% 	\begin{threeparttable} 
	% 		\begin{tabular}{@{}cccccccc@{}}
	% 			\hline
	% 			Method & ISUP & Acc & Prec & Recall & F2-score & AUC  \\ 
	% 			\hline
	% 			Ours & 2-3vs4-5 & 0.800  & 0.471  & 0.810  &0.708    & 0.804    \\ 
	% 			HiFuse & 2-3vs4-5 & 0.789  &0.631  & 0.368  &0.402    & 0.789  \\ 
	% 			MedViT & 2-3vs4-5 & 0.526  &0.5  & 0.315   &0.823    & 0.539  \\ 
	% 			M3T	 & 2-3vs4-5 & 0.750  &0.556  & 0.444   &0.463     & 0.642  \\ 
	% 			Gianluca's & 2-5 & -  &-  & -   &-   & 0.713  \\ 
	% 			\hline
	% 		\end{tabular}
	% 		\begin{tablenotes}    % Start from here   
	% 			\setlength\labelsep{0pt}
	% 			\footnotesize
	% 			\item[1] The isup column indicates the classification task. Prec refer to Precision. M3T \cite{9878673},HiFuse \cite{huo2022hifuse}, MedViT \cite{Manzari_2023} are methods for Comparison. Gianluca's data were taken from the paper \cite{carloni2023causalitydriven}, whose AUC is computed between ISUP 2 vs. rest.
	% 		\end{tablenotes} 
	% 	\end{threeparttable}
	% \end{table}
	\begin{table}[htbp]
 \centering
\caption{Comparison Experiment Results}\label{table:Comparison_experiment}
\begin{tabular}{@{}cccccccc@{}}
\toprule
Method & ISUP & Acc & Prec & Recall & F2-score & AUC  \\ 
\midrule
Ours & 2-3vs4-5 & 0.800  & 0.471  & 0.810  & 0.708 & 0.804 \\ 
HiFuse & 2-3vs4-5 & 0.789 & 0.631 & 0.368 & 0.402 & 0.789 \\ 
MedViT & 2-3vs4-5 & 0.526 & 0.5 & 0.315 & 0.823 & 0.539 \\ 
M3T & 2-3vs4-5 & 0.750 & 0.556 & 0.444 & 0.463 & 0.642 \\ 
Gianluca's & 2-5 & - & - & - & - & 0.713 \\ 
\bottomrule
\end{tabular}
\footnotetext[1]{The ISUP column indicates the classification task. Prec refers to Precision. M3T \cite{9878673}, HiFuse \cite{huo2022hifuse}, MedViT \cite{Manzari_2023} are methods for comparison. Gianluca's data were taken from the paper \cite{carloni2023causalitydriven}, whose AUC is computed between ISUP 2 vs. rest.}
\end{table}

	\begin{figure}[ht]
		\centering
		\includegraphics[width=0.65\linewidth]{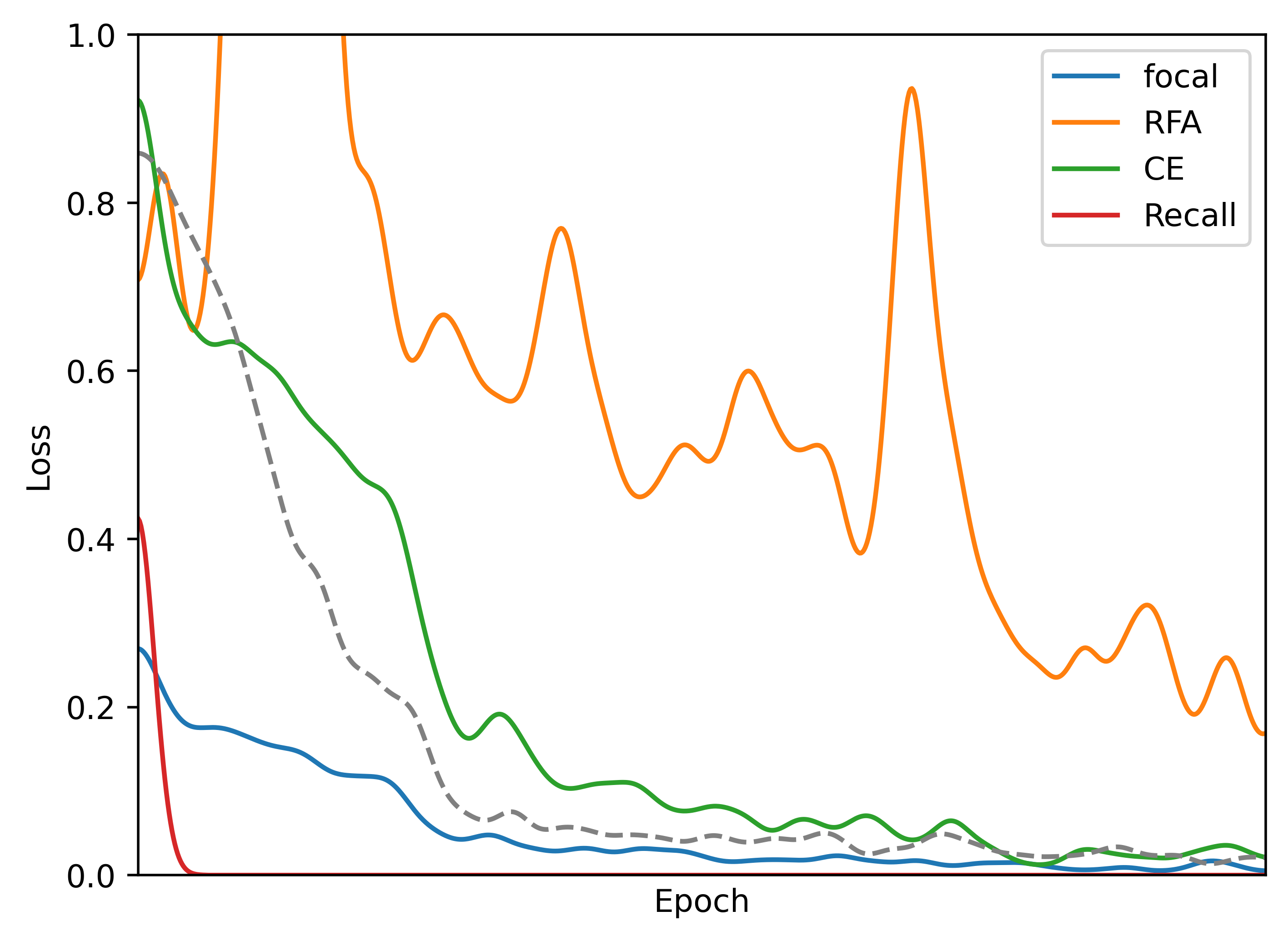}
		\caption{Illustration of the loss function descent during training. In terms of convergence trend, the recall loss quickly converges but gets stuck in a local optimal solution; CE loss and focal loss converge straightforwardly and rapidly. Our proposed RFA loss exhibits significant fluctuations during descent. When feedback is masked in RFAloss (represented by the gray dashed line), it shows rapid convergence without feedback. Therefore, the dynamic feedback mechanism induces fluctuations in the loss function, which helps in searching for optimal parameters.}
		\label{fig:horizontal}
	\end{figure}
	
	%\begin{figure*}[ht]
	%	\centering
	%	\subfloat[$C^1$]
	%	{\includegraphics[width=6cm]{pic/01_2345.png}\label{01_2345}}
	%	% \hspace{-1mm}
	%	\subfloat[$C^2$]
	%	{\includegraphics[width=6cm]{pic/23_45.png}\label{23_45}}
	%	% \hspace{-13mm}
	%	\subfloat[$C^3$]
	%	{\includegraphics[width=6cm]{pic/45.png}\label{45}}\\
	%	\subfloat[ours]
	%	{\includegraphics[width=8cm]{pic/级联查全.png}\label{cascade_recall}}
	%	\subfloat[baseline]
	%	{\includegraphics[width=8cm]{pic/多分类查全.png}\label{multiclass_recall}}
	%	\caption{Figures \ref{01_2345}, \ref{23_45}, and \ref{45} show the classification results on the validation set using optimal parameters for classifiers $C^1$, $C^2$, and $C^3$ respectively. These classifiers were trained to distinguish between ISUP 0-1 vs. 2-5, ISUP 2-3 vs. 4-5, and ISUP 4 vs. 5. It can be observed that despite severely imbalanced class distributions, the classification results exhibit a diagonal pattern with consistently high recall rates.Figure \ref{cascade_recall} represents the recall confusion matrix computed from Figures \ref{01_2345}, \ref{23_45}, and \ref{45}. Figure \ref{multiclass_recall} shows the recall confusion matrix obtained from traditional six-class classification for ISUP using cross-entropy loss.}
	%	\label{fig:cascade}
	%\end{figure*}
	\begin{figure}[htbp]
		\centering
		\includegraphics[width=1\linewidth]{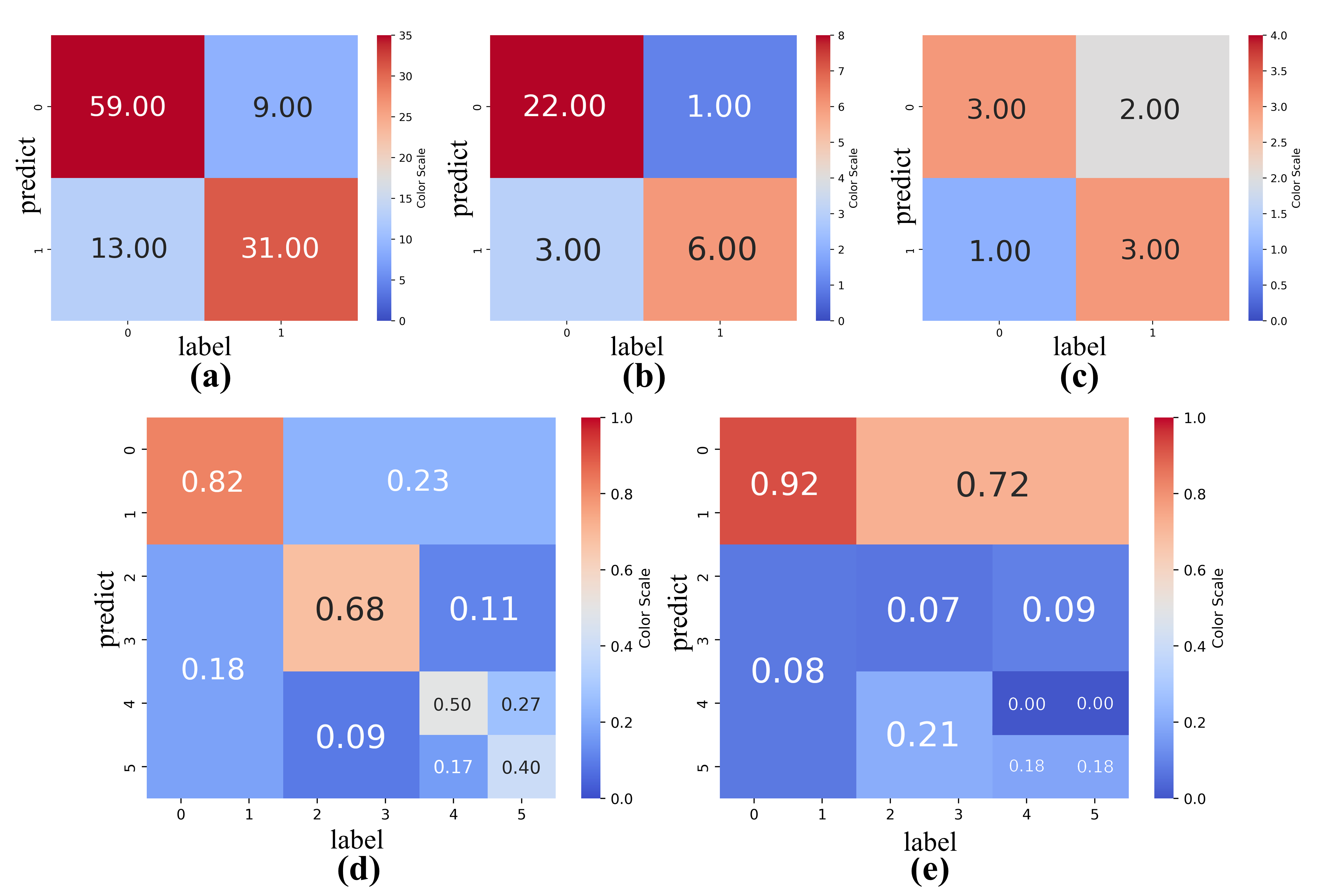}
		\caption{Figures (a), (b), and (c) show the classification results on the validation set using optimal parameters for classifiers $C^1$, $C^2$, and $C^3$ respectively. It can be observed that despite severely imbalanced class distributions, the classification results exhibit a diagonal pattern with consistently high recall rates.Figure (d) represents the recall confusion matrix computed from Figures (a), (b), and (c). Figure (e) shows the recall confusion matrix obtained from traditional six-class classification for ISUP using cross-entropy loss.}
		\label{fig:muti}
	\end{figure}
	
	We first experimented with the hyperparameters of the RFA loss function. The experimental results are shown in Table \ref{table:control_variables}, and the trend of loss reduction is depicted in Figure \ref{pic:control parameter curve}. The performance of experiments $n_1$ and $n_2$ initially improves but then declines as the variables change progressively. According to the results of parameter $M$, the recall, ARS, and F2-score got the best result when $M$ was set to 0.3. However, as $M$ increases, the values of evaluation metrics noticeably decrease. And the increasing of three hyperparameters leads to greater fluctuations in the entire curve. Conversely, when $n_1$ is small, changes in it have less impact. The experimental results roughly demonstrate the effects of the hyperparameters of the RFA loss function. However, we observed that the degree of fluctuation does not necessarily impact the final scores, suggesting only a limited correlation. We conducted a detailed analysis of this issue, which will be discussed in the DISCUSSION section.

	Table \ref{table:ablation_experiment} summarizes the results of the ablation experiment, with F-E representing feature extraction. We used the cross-entropy loss function as our baseline of ablation experiment. When we applied the recall loss function, it predicted all samples as category 1, resulting in a recall rate of 1. When we used the RFA loss, the F2-score, AUC had a certain improvement compared to the other loss; In terms of the recall rate, there was an improvement of 16.2\% compared to cross-entropy and 29.2\% compared to focal loss functions.  Applying feature extraction improved the F2-score by 20.6\% ,and the AUC by 13.5\% for the Focal loss, while the experimental results of cross-entropy loss decreased. Applying both feature extraction and RFA loss simultaneously improves the F2-score by 7.9\%, AUC by 4.8\%, and recall rate by 12.9\% compared to applying RFA alone. Additionally, compared to the baseline, recall rate improves by 29.1\%, F2-score improves by 20.5\%, and AUC improves by 10.9\%. These results demonstrate that our work significantly improves the recall rate while achieving excellent accuracy.
	
	Table \ref{table:Comparison_experiment} shows the experimental results of the comparative experiment. The table shows that our method has significantly improved the recall rate, 36.6\% higher than the second place, while maintaining the high accuracy and AUC.

	We performed experiments on the cascaded refinement strategy using the optimal parameter set.We also used the cross-entropy loss function as our baseline of cascaded experiment. Figures 6a, 6b, and 6c demonstrate that the confusion matrices have a diagonal distribution and maintain high recall rates.Figure 6d is the recall confusion matrix calculated from Figure 6a, 6b, and 6c; Figure 6e is the recall confusion matrix derived from traditional six-class classification for ISUP task using cross-entropy loss function. Compared to the baseline, our cascaded refinement strategy renders the overall results more balanced and focuses more on the diagonal positions. The recall rate for each category decreases as the ISUP grade increases. However, even for the most challenging category five, there is still a 40\% recall rate. This result supports the effectiveness of our data processing strategy. 
	
	\section{DISCUSSION}
	
	\subsection{Practical Significance of RFAloss Hyperparameters}
	
	We will discuss the effect and interpretable hyperparameters tuning strategies below. $n_1$, $n_2$, and $M$ are adjustable hyperparameters that affect the accuracy ($a$), recall rate ($r$), and equation \ref{eq:A0part}, respectively. The magnitude of the feedback effect can be reflected by the fluctuation of the curves in Figure \ref{pic:control parameter curve}, and the final results are shown in Table \ref{table:control_variables}.
	
	Based on the results and our original design intent, $n_2$ aligns with our hypothesis and has significant effects. It directly affects the recall rate, resulting in an exponential growth in the impact of the recall. This leads to an increase in $\mathcal A$ and a greater focus on positive examples over negative ones. The results also validate our prediction, as shown in Figure \ref{n2pic}. Increasing $n_2$ noticeably increases the amplitude of fluctuations, indicating a stronger penalty on dynamic recall for each update. However, the experiments in Table \ref{table:control_variables} show that increased fluctuation indicates stronger feedback and a wider search range but does not necessarily lead to improved final metric results.
	
	We introduce $n_1$ to improve accuracy by directing feedback towards accuracy improvement. In our study, the positive class is significantly underrepresented. Additionally, our loss function is designed to prioritize a high recall rate by predicting more positive examples. Therefore, we attribute the lower accuracy to the scarcity of predictions for negative samples. To improve the accuracy, we need to increase the attention to negative examples. When $n_1<0$, the increase in $a^{n_1}$ significantly enhances the effect, causing the feedback $1-a^{n_1}$ to be smaller than $1-a$. This indicates that there is a higher attention to negative examples compared to positive examples, which supports our hypothesis. However, if $n_1$ is too small, $a^{n_1}$ tends to approach 1. As a result, there is an insufficient fluctuation effect that significantly weakens the search capability during training. This explains why the curve for $n_1=1$ in Figure \ref{n1pic} has a significantly larger fluctuation amplitude compared to other values. In contrast, the overall fluctuation for $n_1=0.25$ and $n_1=0.5$ is not significant. When $n_1=0.75$, the effect is optimal because it retains the sensitive feedback capabilities and better balances the relationship between accuracy and recall. 
	
	The above description explains the individual properties of $n_1$ and $n_2$. When they work together, they have a combined effect on equation\ref{eq:A0part}. To make RFAloss controllable, we introduce parameter $M$ as an auxiliary control that can modify the baseline of equation\ref{eq:Apart}. If the penalty of $n_2$ on the recall rate is too large, leading to excessive attention to positive examples, we can decrease $M$ to artificially reduce the weight of positive examples.Based on this, Figure \ref{Cpic} can be interpreted as follows: $n_2=3$ represents a parameter setting that prioritizes recall rate. The curve has greater fluctuation with a high baseline when $M=0.7$ compared to $M=0.3$ and $M=0.5$. The curve at $M=0.3$ fluctuates but tends towards stability. The corresponding indicators in Table \ref{table:control_variables} also show that $M=0.3$ yields the optimal results in the parameter $M$ experiments.
	
	Finally, we will discuss the relationship between curve fluctuations and metric results, as well as the issue of parameter selection. RFAloss incorporates validation set accuracy and recall into training process, with hyperparameters describing the sensitivity of parameter $\mathcal{A}$ to feedback. When recall is low, the loss function amplifies attention to positives, penalizing false negatives more significantly, prompting a shift in the loss function's benchmark and search direction. It ensures the discovery of more samples with outstanding recall. Therefore, the amplitude of curve fluctuations should not be too large to maintain correct search direction. Hence, we need to consider the interpretability of feedback hyperparameters and select a parameter set with excellent feedback sensitivity and precise feedback direction: $n_1$ should not be too small, $n_2$ can be adjusted according to the requirement for recall attention, and finally, parameter $M$ is determined.
	
	\subsection{Discussion on Interpretability}
	
	As shown in Figure 6e, the model lacks sensitivity to samples with higher ISUP grades due to two reasons. First, the uneven distribution is caused by disease prevalence. Second, the classification of prostate cancer grades in clinical lacks clear boundaries, making it challenging to train a comprehensive understanding model due to the complexity of ambiguous medical knowledge involved. In clinical practice, diagnosis also follows a stepwise grading approach for clinical decision-making. Our cascaded refinement strategy models this process: Classifier $C^1$ trains the model for diagnosing csPCa; classifiers $C^2$ and $C^3$ differentiate between different degrees of disease severity, reflecting varying levels of clinical intervention. 
	
	Our research focuses on the recall rate in medical tasks. We prioritize modeling the clinical decision-making aspect of Classifier $C^2$ as it is pivotal. The cascade training directly incorporates optimal hyperparameters from $C^2$, resulting in outstanding results that demonstrate the superior generality of our loss function.
	
	Lastly, our work consistently aims to assist physicians. Our feature extraction maps can serve as diagnostic aids for doctors, while the cascaded refinement strategy can provide flexible confidence levels based on mpMRI. Practitioners can use individual classifiers for specific clinical applications, enhancing their diagnostic capabilities.

	\section{CONCLUSIONS}
	We propose a recall-guided deep learning-assisted ISUP grading strategy based on mpMRI. Compared to the baseline, our approach improves recall by 29.1\% . Our work emphasizes the practical significance by integrating ISUP grading indicators and diagnostic processes of prostate cancer into deep learning. Our primary contribution lies in introducing a universal Recall Feedback Adaptive loss function that prioritizes low prevalence and low quantity labels. This loss function enhances the search direction and scope during the training process. Furthermore, our prior knowledge-based feature extraction strategy amplifies the differences between lesion areas and their surroundings, providing prior information to the model. Under the premise of RFAloss, this approach increases recall by 12.9\% and the accuracy is maintained. We implement a cascaded refinement strategy, which results in a diagonal confusion matrix for the recall metric. These methods are valuable references for medical image processing and its practical applications.

\bibliographystyle{unsrt}

% \bibliography{sn-bibliography}% common bib file
% %% if required, the content of .bbl file can be included here once bbl is generated
% %%\input sn-article.bbl
\end{document}